\documentclass[11pt,a4paper]{article}
\usepackage[hyperref]{acl2019}
\usepackage[utf8]{inputenc}
\usepackage{times}
\usepackage{latexsym}
\usepackage{graphicx}
\usepackage{arydshln}
\usepackage{verbatim}

\usepackage{url}
\usepackage{multirow}
\usepackage{booktabs}
\usepackage{pgfplots}

\usepackage{expex}
\lingset{everygl=\small, extraglskip=!.75ex, belowglpreambleskip=-0.5ex, aboveglftskip=-0.5ex}
\aclfinalcopy 

\title{The University of Helsinki submissions to the\\ WMT19 news translation task}

\author{Aarne Talman,\textsuperscript{$\ast$ $\dagger$} Umut Sulubacak,\textsuperscript{$\ast$} Ra\'ul V\'azquez,\textsuperscript{$\ast$} Yves Scherrer,\textsuperscript{$\ast$} Sami Virpioja,\textsuperscript{$\ast$} \\{\bf Alessandro Raganato},\textsuperscript{$\ast$ $\dagger$} {\bf Arvi Hurskainen}\textsuperscript{$\ast$} and {\bf J\"org Tiedemann}\textsuperscript{$\ast$} \\~\\
\textsuperscript{$\ast$}University of Helsinki\\
\textsuperscript{$\dagger$}Basement AI \\
\texttt{\{name.surname\}@helsinki.fi}}

\date{}

\begin{document}
\maketitle

\begin{abstract}
In this paper, we present the University of Helsinki submissions to the WMT 2019 shared task on news translation in three language pairs: English--German, English--Finnish and Finnish--English. This year, we focused first on cleaning and filtering the training data using multiple data-filtering approaches, resulting in much smaller and cleaner training sets. For English--German, we trained both sentence-level transformer models and compared different document-level translation approaches. For Finnish--English and English--Finnish we focused on different segmentation approaches, and we also included a rule-based system for English--Finnish.
\end{abstract}

\section{Introduction}

The University of Helsinki participated in the WMT 2019 news translation task with four primary submissions. We submitted neural machine translation systems for English-to-Finnish, Finnish-to-English and English-to-German, and a rule-based machine translation system for English-to-Finnish.

Most of our efforts for this year's WMT focused on data selection and pre-processing (Section~\ref{sec:data}), sentence-level translation models for English-to-German, English-to-Finnish and Finnish-to-English (Section~\ref{sec:sent}), document-level translation models for English-to-German (Section~\ref{sec:doclevel}), and a comparison of different word segmentation approaches for Finnish (Section~\ref{sec:segm}). The final submitted NMT systems are summarized in Section~\ref{sec:finalsystems}, while the rule-based machine translation system is described in Section~\ref{sec:rbmt}.

\section{Pre-processing, data filtering and back-translation} \label{sec:data}
It is well known that data pre-processing and selection has a huge effect on translation quality in neural machine translation. We spent substantial effort on filtering data in order to reduce noise---especially in the web-crawled data sets---and to match the target domain of news data.

The resulting training sets, after applying the steps described below, are for 15.7M sentence pairs for English--German, 8.5M sentence pairs for English--Finnish, and 12.3M--26.7M sentence pairs (different samplings of back-translations) for Finnish--English.

\subsection{Pre-processing}
For each language, we applied a series of pre-processing steps using scripts available in the Moses decoder~\cite{moses_dec}:
\begin{itemize}
\setlength\itemsep{0pt}
    \item replacing unicode punctuation,
    \item removing non-printing characters,
    \item normalizing punctuation,
    \item tokenization.
\end{itemize}
In addition to these steps, we replaced a number of English contractions with the full form, \emph{e.g.} \textit{``They're''}\,$\rightarrow$\,\textit{``They are''}.
After the above steps, we applied a Moses truecaser model trained for individual languages, and finally a byte-pair encoding~(BPE)~\cite{sennrich-etal-2016-bpe} segmentation using a set of codes for either language pair.

For English--German, we initially pre-processed the data using only punctuation normalization and tokenization. We subsequently trained an English truecaser model using all monolingual English data as well as the English side of all parallel English--German datasets except the Rapid corpus (in which non-English characters were missing from a substantial portion of the German sentences). We also repeated the same for German. Afterwards, we used a heuristic cleanup script\footnote{Shared by Marcin Junczys-Dowmunt. Retrieved from \url{https://gist.github.com/emjotde/4c5303e3b2fc501745ae016a8d1e8e49}} in order to filter suspicious samples out of Rapid, and then truecased all parallel English--German data (including the filtered Rapid) using these models. Finally, we trained BPE codes with 35\,000 symbols jointly for English--German on the truecased parallel sets. For all further experiments with English--German data, we applied the full set of tokenization steps as well as truecasing and BPE segmentation.

For English--Finnish, we first applied the standard tokenization pipeline. For English and Finnish respectively, we trained truecaser models on all English and Finnish monolingual data as well as the English and Finnish side of all parallel English--Finnish datasets. As we had found to be optimal in our previous year submission~\cite{raganato-etal-2018}, we trained a BPE model using a vocabulary of 37\,000 symbols, trained jointly only on the parallel data. Furthermore, for some experiments, we also used domain labeling. We marked the datasets with 3 different labels: \textit{$\langle$NEWS$\rangle$} for the development and test data from 2015, 2016, 2017, \textit{$\langle$EP$\rangle$} for Europarl, and \textit{$\langle$WEB$\rangle$} for ParaCrawl and Wikititles.

%
%
%

\subsection{Data filtering}
 For data filtering we applied four types of filters: (i) rule-based heuristics, (ii) filters based on language identification, (iii) filters based on word alignment models, and (iv) language model filters.

\paragraph{Heuristic filters:}

The first step in cleaning the data refers to a number of heuristics (largely inspired by \cite{stahlberg-degispert-byrne:2018:WMT}) including:
\begin{itemize}
    \item removing all sentence pairs with a length difference ratio above a certain threshold: for CommonCrawl, ParaCrawl and Rapid we used a threshold of 3, for WikiTitles a threshold of 2, and for all other data sets a threshold of 9;
    \item removing pairs with short sentences: for CommonCrawl, ParaCrawl and Rapid we required a minimum number of four words;
    \item removing pairs with very long sentences: we restricted all data to a maximum length of 100 words;
    \item removing sentences with extremely long words: We excluded all sentence pairs with words of 40 or more characters;
    \item removing sentence pairs that include HTML or XML tags;
    \item decoding common HTML/XML entities;
    \item removing empty alignments (while keeping document boundaries intact);
    \item removing pairs where the sequences of non-zero digits occurring in either sentence do not match;
    \item removing pairs where one sentence is terminated with a punctuation mark and the other is either missing terminal punctuation or terminated with another punctuation mark.
\end{itemize}


\paragraph{Language identifiers:} There is a surprisingly large amount of text segments in a wrong language in the provided parallel training data. This is especially true for the ParaCrawl and Rapid data sets. This is rather unexpected as a basic language identifier certainly must be part of the crawling and extraction pipeline. Nevertheless, after some random inspection of the data, we found it necessary to apply off-the-shelf language identifiers to the data for removing additional erroneous text from the training data. In particular, we applied the Compact Language Detector version 2 (CLD2) from the Google Chrome project (using the Python interface from \textit{pycld2}\footnote{\url{https://github.com/aboSamoor/pycld2}}), and the widely used \textit{langid.py} package \cite{lui-baldwin-2012-langid} to classify each sentence in the ParaCrawl, CommonCrawl, Rapid and Wikititles data sets. We removed all sentence pairs in which the language of one of the aligned sentences was not reliably detected. For this, we required the correct language ID from both classifiers, the reliable-flag set to ``True'' by CLD2 with a reliability score of 90 or more, and the detection probability of \textit{langid.py} to be at least 0.9.
%
\paragraph{Word alignment filter:} Statistical word alignment models implement a way of measuring the likelihood of parallel sentences. IBM-style alignment models estimate the probability $p(f\,|\,a,e)$ of a foreign sentence $f$ given an "emitted" sentence $e$ and an alignment $a$ between them. Training word alignment models and aligning large corpora is very expensive using traditional methods and implementations. Fortunately, we can rely on {\em eflomal}\footnote{Software available from \url{https://github.com/robertostling/eflomal}}, an efficient word aligner based on Gibbs sampling \cite{Ostling2016efmaral}. Recently, the software has been updated to allow the storage of model priors that makes it possible to initialize the aligner with previously stored model parameters. This is handy for our filtering needs as we can now train a model on clean parallel data and apply that model to estimate alignment probabilities of noisy data sets. 

We train the alignment model on Europarl and news test sets from previous WMTs for English--Finnish, and NewsCommentary for English--German. For both language pairs, we train a Bayesian HMM alignment model with fertilities in both directions and estimate the model priors from the symmetrized alignment. We then use those priors to run the alignment of the noisy data sets using only a single iteration of the final model to avoid a strong influence of the noisy data on alignment parameters. As it is intractable to estimate a fully normalized conditional probability of a sentence pair under the given higher-level word alignment model, eflomal estimates a score based on the maximum unnormalized log-probability of links in the last sampling iteration. In practice, this seems to work well, and we take that value to rank sentence pairs by their alignment quality. In our experiments, we set an arbitrary threshold of 7 for that score, which seems to balance recall and precision well according to some superficial inspection of the ranked data. The word alignment filter is applied to all web data as well as to the back-translations of monolingual news.

\paragraph{Language model filter:}
The most traditional data filtering method is probably to apply a language model. The advantage of language models is that they can be estimated from monolingual data, which may be available in sufficient amounts even for the target domain. In our approach, we opted for a combination of source and target language models and focused on the comparison between scores coming from both models. The idea is to prefer sentence pairs for which not only the cross-entropy of the individual sentences ($H(S,q_s)$ and $H(T,q_t)$) is low with respect to in-domain LMs, but also the absolute difference between the cross-entropies ($abs(H(S,q_s)-H(T,q_t))$) for aligned source and target sentences is low. The intuition is that both models should be roughly similarly surprised when observing sentences that are translations of each other.
In order to make the values comparable, we trained our language models on parallel data sets. 

For English--Finnish, we used news test data from 2015-2017 as the only available in-domain parallel training data, and for English--German we added the NewsCommentary data set to the news test sets from 2008-2018. As both data sets are small, and we aimed for an efficient and cheap filter, we opted for a traditional n-gram language model in our experiments. To further avoid data sparseness and to improve comparability between source and target language, we also based our language models on BPE-segmented texts using the same BPE codes as for the rest of the training data. {\em VariKN}~\cite{journals/taslp/SiivolaHV07,d74e7d677f0b4358958795403e5e10a7}\footnote{VariKN is available from \url{https://vsiivola.github.io/variKN/}} is the perfect toolkit for the purposes of estimating n-gram language models with subword units. It implements Kneser-Ney growing and revised Kneser-Ney pruning methods with the support of n-grams of varying size and the estimation of word likelihoods from text segmented in subword units. In our case, we set the maximum n-gram size to 20, and the pruning threshold to 0.002. Finally, we computed cross-entropies for each sentence in the noisy parallel training data and stored 5 values as potential features for filtering: $H(S,q_s)$, $H(T,q_t)$, $avg(H(S,q_s),H(T,q_t))$, $max(H(S,q_s),H(T,q_t))$ and $abs(H(S,q_s)-(T,q_t))$. Based on some random inspection, we selected a threshold of 13 for the average cross-entropy score, and a threshold of 4 for the cross-entropy difference score. For English--Finnish, we opted for a slightly more relaxed setup to increase coverage, and set the average cross-entropy to 15 and the difference threshold to 5. We applied the language model filter to all web data and to the back-translations of monolingual news.

\begin{table}[h!]
    \centering
    \begin{tabular}{lcc}
    \toprule
                & EN--DE    & EN--FI\\
    \midrule
CommonCrawl     & 3.2\%     & \\
Europarl        & 0.8\%     & 2.8\% \\
News-Commentary & 0.2\%     & \\
ParaCrawl       & 0.6\%     & \\
Rapid           & 13.2\%    & 5.2\% \\
WikiTitles      & 8.0\%     & 4.0\% \\
\bottomrule
    \end{tabular}
    \caption{Basic heuristics for filtering -- percentage of lines removed. For English--Finnish the statistics for ParaCrawl are not available because the cleanup script was applied after other filters.}
    \label{tab:basic-cleanup}
\end{table}

\paragraph{Applying the filter to WMT 2019 data:} The impact of our filters on the data provided by WMT 2019 is summarized in Tables~\ref{tab:basic-cleanup},~\ref{tab:filter-ende} and~\ref{tab:filter-enfi}.

\begin{table}[h!]
    \centering
    \begin{tabular}{lrrr}
    \toprule
                & \multicolumn{3}{c}{\% rejected} \\
    \cmidrule(lr){2-4}
    Filter      & CC        & ParaCrawl    & Rapid \\
\midrule
LM average CE   & 31.9\%    & 62.0\%    & 12.7\% \\
LM CE diff      & 19.0\%    & 12.7\%    &  6.9\% \\
Source lang ID  &  4.0\%    & 30.7\%    &  7.3\% \\
Target lang ID  &  8.0\%    & 22.7\%    &  6.2\% \\
Wordalign       & 46.4\%    &  3.1\%    &  8.4\% \\
Number          & 15.3\%    & 16.0\%    &  5.0\% \\
Punct           &  0.0\%    & 47.4\%    & 18.7\% \\
\midrule
total           & 66.7\%    & 74.7\%    & 35.1\% \\
\bottomrule
    \end{tabular}
    \caption{Percentage of lines rejected by each filter for English--German data sets. Each line can be rejected by several filters. The total of rejected lines is the last row of the table.}
    \label{tab:filter-ende}
\end{table}

\begin{table}[h!]
    \centering
    \begin{tabular}{lrrrr}
    \toprule
    & \multicolumn{4}{c}{\% rejected} \\
    \cmidrule(lr){2-5}
    & \multicolumn{2}{c}{ParaCrawl} & \multicolumn{2}{c}{Rapid} \\
    \cmidrule(lr){2-3} \cmidrule(lr){4-5}
    Filter            & strict    & relax      & strict    & relax \\
\midrule
LM avg CE       & 62.5\%    & 40.0\%    & 50.7\% & 21.4\% \\
LM CE diff      & 35.4\%    & 25.7\%    & 44.8\% & 31.1\% \\
Src lang ID     & 37.2\%    & 37.2\%    & 11.9\% & 11.9\% \\
Trg lang ID     & 29.1\%    & 29.1\%    &  8.5\% &  8.5\% \\
Wordalign       &  8.3\%    &  8.3\%    &  8.3\% &  8.3\% \\
Number          & 16.8\%    & 16.8\%    &  6.7\% &  6.7\% \\
Punct           & 54.6\%    &  3.3\%    & 23.7\% &  7.6\% \\
\midrule
total           & 87.9\%    & 64.2\%    & 62.2\% & 54.8\% \\
\bottomrule
    \end{tabular}
    \caption{Percentage of lines rejected by each filter for English--Finnish data sets. The strict version is the same as for English--German, and the relax version applies relaxed thresholds.}
    \label{tab:filter-enfi}
\end{table}

We can see that the ParaCrawl corpus is the one that is the most affected by the filters. A lot of noise can be removed, especially by the language model filter. The strict punctuation filter also has a strong impact on that data set. Naturally, web data does not come with proper complete sentences that end with proper final punctuation marks, and the filter might remove quite a bit of the useful data examples. However, our final translation scores reflect that we do not seem to lose substantial amounts of performance even with the strict filters. Nevertheless, for English--Finnish, we still opted for a more relaxed setup to increase coverage, as the strict version removed over 87\% of the ParaCrawl data.

It is also interesting to note the differences of individual filters on different data sets. The word alignment filter seems to reject a large portion of the CommonCrawl data set whereas it does not affect other data sets that much. The importance of language identification can be seen with the ParaCrawl data whereas other corpora seem to be much cleaner with respect to language.

\subsection{Back-translation}

We furthermore created synthetic training data by back-translating news data. We translated the monolingual English news data from the years 2007--2018, from which we used a filtered and sampled subset of 7M sentences for our Finnish--English systems, and the Finnish data from years 2014--2018 using our WMT 2018 submissions. We also used the back-translations we generated for the WMT 2017 news translation task, where we used an SMT model to create 5.5M sentences of back-translated data from the Finnish news2014 and news2016 corpora \cite{ostling-etal-2017-helsinki}.

For the English--German back-translations, we trained a standard transformer model on all the available parallel data and translated the monolingual German data into English. The BLEU score for our back-translation model is 44.24 on news-test 2018. We applied our filtering pipeline to the back-translated pairs, resulting in 10.3M sentence pairs. In addition to the new back-translations, we also included back-translations from the WMT16 data by~\citet{sennrich2016improving}.

\section{Sentence-level approaches} \label{sec:sent}

In this section we describe our sentence-level translation models and the experiments in the English-to-German, English-to-Finnish and Finnish-to-English translation directions.

\subsection{Model architectures}
We experimented with both NMT and rule-based systems. All of our neural sentence-level models are based on the transformer architecture~\citep{transformer}. We used both the OpenNMT-py~\citep{opennmt} and MarianNMT~\citep{mariannmt} frameworks. Our experiments focused on the following:
\begin{itemize}
    \item Ensemble models: using ensembles with a combination of independent runs and save-points from a single training run.
    \item Left-to-right and right-to-left models: Transformer models with decoding of the output in left-to-right and right-to-left order.
\end{itemize}

The English-to-Finnish rule-based system is an enhanced version of the WMT 2018 rule-based system~\citep{raganato-etal-2018}.

\subsection{English--German}  \label{sec:en_de}


Our sentence-level models for the English-to-German direction are based on ensembles of independent runs and different save-points as well as save-points fine-tuned on in-domain data. For our submission, we used an ensemble of 9 models containing:
\begin{itemize}
    \item 4 save-points with the lowest development perplexity taken from a model trained for 300\,000 training steps.
    \item 5 independent models fine-tuned with in-domain data.
\end{itemize}
All  our sentence-level models for the English--German language pair are  trained  on  filtered  versions of  Europarl, NewsCommentary, Rapid, CommonCrawl, ParaCrawl, Wikititles, and back-translations. For in-domain fine-tuning, we use newstest 2011--2016. Our submission is composed of transformer-big models implemented in OpenNMT-py with 6 layers of hidden size 4096, 16 attention heads, and a dropout of 0.1. The differences in development performance between the best single model, an ensemble of save-points of a single training run and our final submission are reported in Table~\ref{tab:opennmt_ende}. We gain 2 BLEU points with the ensemble of save-points, and an additional 0.8 points by adding in-domain fine-tuned models into the ensemble. This highlights the well-known effectiveness of ensembling and domain adaptation for translation quality.

\begin{table}[h!]
    \centering
    \begin{tabular}{lc}
    \toprule
            & BLEU news2018 \\
    \midrule
    Single model   & 44.61 \\
    5 save-points  & 46.65 \\
    5 save-points + 4 fine-tuned  & \bf 47.45 \\
    \bottomrule
    \end{tabular}
    \caption{English--German development results comparing the best single model, an ensemble of 5 save-points, and an ensemble of 5 save-points and 4 independent runs fine-tuned on in-domain data.}
    \label{tab:opennmt_ende}
\end{table}

Furthermore, we trained additional models using MarianNMT with the same training data and fine-tuning method. In this case, we also included right-to-left decoders that are used as a complement in the standard left-to-right decoders in re-scoring approaches. In total, we also end up with 9 models including:

\begin{itemize}
\setlength\itemsep{0pt}
    \item 3 independent models trained for left-to-right decoding,
    \item 3 independent models trained for right-to-left decoding,
    \item 3 save-points based on continued training of one of the left-to-right decoding models.
\end{itemize}

The save-points were added later as we found out that models kept on improving when using larger mini-batches and less frequent validation in early stopping. Table~\ref{tab:marian_ende} lists the results of various models on the development test data from 2018.

\begin{table}[h!]
    \centering
    \begin{tabular}{lcc}
    \toprule
            & \multicolumn{2}{c}{BLEU news2018} \\
    Model   & Basic & Fine-tuned \\
    \midrule
    L2R run 1    & 43.63 & 45.31 \\
    L2R run 2    & 43.52 & 45.14 \\
    L2R run 3    & 43.33 & 44.93 \\
    L2R run3 cont'd 1 & 43.65 & 45.11\\
    L2R run3 cont'd 2 & 43.76 & 45.43\\
    L2R run3 cont'd 3 & 43.53 & 45.67\\
    \midrule
    Ensemble all L2R  & 44.61 & 46.34\\
    Rescore all L2R   &       & 46.49\\
    \midrule
    \midrule
    R2L run 1    & 42.14 & 43.80 \\
    R2L run 2    & 41.96 & 43.67 \\
    R2L run 3    & 42.17 & 43.91 \\
    \midrule
    Ensemble all R2L & 43.03 & 44.70 \\
    Rescore all R2L  &       & 44.73 \\
    \midrule
    \midrule
    Rescore all L2R+R2L &    & {\bf 46.98}\\
    \bottomrule
    \end{tabular}
    \caption{English--German results from individual MarianNMT transformer models and their combinations (cased BLEU).}
    \label{tab:marian_ende}
\end{table}

There are various trends that are interesting to point out. First of all, fine-tuning gives a consistent boost of 1.5 or more BLEU points. Our initial runs were using a validation frequency of 5\,000 steps and a single GPU with dynamic mini-batches that fit in 13G of memory. The stopping criterion was set to 10 validation steps without improving cross-entropy on heldout data (newstest 2015 + 2016). Later on, we switched to multi-GPU training with two GPUs and early stopping of 20 validation steps. The dynamic batching method of MarianNMT produces larger minibatches once there is more memory available, and multi-GPU settings simply multiply the working memory for that purpose. We realized that this change enabled the system to continue training substantially, and Table~\ref{tab:marian_ende} illustrates the gains of that process for the third L2R model.

Another observation is that right-to-left decoding models in general work less well compared to the corresponding left-to-right models. This is also apparent with the fine-tuned and ensemble models that combine independent runs. The difference is significant with about 1.5 BLEU points or more. Nevertheless, they still contribute to the overall best score when re-scoring n-best lists from all models in both decoding directions. In this example, re-scoring is done by simply summing individual scores. 
Table~\ref{tab:marian_ende} also shows that re-scoring is better than ensembles for model combinations with the same decoding direction because they effectively increase the beam size as the hypotheses from different models are merged before re-ranking the combined and re-scored n-best lists.

\begin{figure}[h!]
	\centering
 	\begin{tikzpicture}
 	\definecolor{color1}{RGB}{15,78,156}
 	\definecolor{color2}{RGB}{94,139,31}
 	\begin{axis}[
     	height=6.0cm, width=\linewidth,
 		xlabel=Beam size,
 		ylabel=BLEU news 2018,
 		xmode=log,
 		xtick=data,
 		xticklabels={1,2,4,8,16,32,64},
 		ytick distance=0.5,
 		yticklabels={,,,43.0,43.5,44.0,44.5,45.0,45.5,46.0,46.5},
 		legend pos=south east,
 		ymajorgrids
 		]
 	\addplot[color1, mark=*, line width=.4mm] table[x=Beam,y=L2R] {
 Beam    L2R R2L
 1   45.54   42.84
 2   46.22   43.80
 4   46.48   44.31
 8   46.30   44.57
 16  46.36   44.74
 32  46.44   44.78
 64  46.45   44.86
 	};
 	\addplot[color2, mark=diamond*, mark size=2.3pt, line width=.4mm] table[x=Beam,y=R2L] {
 Beam    L2R R2L
 1   45.54   42.84
 2   46.22   43.80
 4   46.48   44.31
 8   46.30   44.57
 16  46.36   44.74
 32  46.44   44.78
 64  46.45   44.86
 	};
 	\legend{L2R, R2L}
 	\end{axis}
 \end{tikzpicture}
	\caption{The effect of beam size on translation performance. All results use model ensembles and the scores are case-sensitive.}
	\label{fig:marian_ende_beam}
\end{figure}
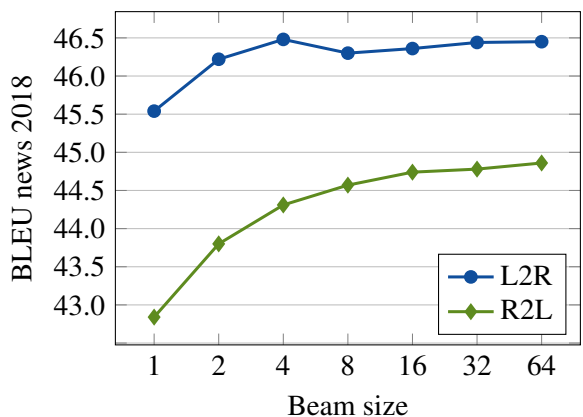

The positive effect of beam search is further illustrated in Figure~\ref{fig:marian_ende_beam}. All previous models were run with a beam size of 12. As we can see, the general trend is that larger beams lead to improved performance, at least until the limit of 64 in our experiments. Beam size 4 is an exception in the left-to-right models.

\subsection{English--Finnish and Finnish--English} \label{sec:segm}

The problem of open-vocabulary translation is particularly acute for morphologically rich languages like Finnish. In recent NMT research, the standard approach consists of applying a word segmentation algorithm such as BPE~\cite{sennrich-etal-2016-bpe} or SentencePiece~\cite{kudo-richardson-2018-sentencepiece} during pre-processing. 
In recent WMT editions, various alternative segmentation approaches were examined for Finnish: hybrid models that back off to character-level representations~\cite{ostling-etal-2017-helsinki}, and variants of the Morfessor unsupervised morphology algorithm~\cite{gronroos-etal-2018-cognate}. This year, we experimented with rule-based word segmentation based on Omorfi~\cite{pirinen-2015-omorfi}. Omorfi is a morphological analyzer for Finnish with a large-coverage lexicon. Its segmentation tool\footnote{\url{https://flammie.github.io/omorfi/pages/usage-examples.html\#morphological-segmentation}} splits a word form into morphemes as defined by the morphological rules. In particular, it distinguishes prefixes, infixes and suffixes through different segmentation markers:

\exdisplay
\begingl
\gla Intia$\rightarrow$ $\leftarrow$n ja Japani$\rightarrow$ $\leftarrow$n pää$\rightarrow$ $\leftarrow$ministeri$\rightarrow$ $\leftarrow$t tapaa$\rightarrow$ $\leftarrow$vat Tokio$\rightarrow$ $\leftarrow$ssa //
\glb India \textsc{gen} and Japan \textsc{gen} prime minister \textsc{pl} meet \textsc{3pl} Tokyo \textsc{ine} //
\endgl
\xe


While Omorfi provides word segmentation based on morphological principles, it does not rely on any frequency cues. Therefore, the standard BPE algorithm is run over the Omorfi-segmented text in order to split low-frequency morphemes.

In this experiment, we compare two models for each translation direction:
\begin{itemize}
\item One model segmented with the standard BPE algorithm (joint vocabulary size of 50\,000, vocabulary frequency threshold of 50).
\item One model where the Finnish side is pre-segmented with Omorfi, and both the Omorfi-segmented Finnish side and the English side are segmented with BPE~(same parameters as above).
\end{itemize}

All models are trained on filtered versions of Europarl, ParaCrawl, Rapid, Wikititles, newsdev2015 and newstest2015 as well as back-translations. Following our experiments at WMT 2018~\cite{raganato-etal-2018}, we also use domain labels~(\textit{$\langle$EP$\rangle$} for Europarl, \textit{$\langle$Web$\rangle$} for ParaCrawl, Rapid and Wikititles, and \textit{$\langle$NEWS$\rangle$} for newsdev, newstest and the back-translations). 
We use newstest2016 for validation. All models are trained with MarianNMT, using the standard Transformer architecture.

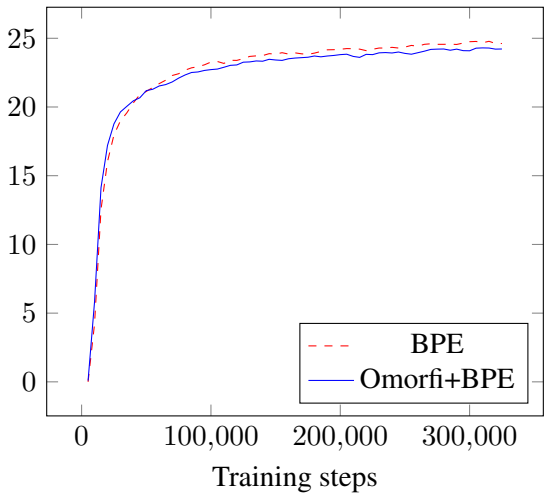
\begin{figure}[h!]
\centering
\begin{tikzpicture}
	\begin{axis}[
	    height=7.0cm,
		xlabel=Training steps,
		x tick label style={/pgf/number format/fixed},
		scaled x ticks=false,
		legend pos=south east
		]
	\addplot[color=red, dashed] table[x=Steps,y=Bleu1] {
Steps	Bleu1
5000	0
10000	4.23
15000	12.55
20000	16.02
25000	17.97
30000	18.99
35000	19.58
40000	20.3
45000	20.9
50000	21.13
55000	21.42
60000	21.7
65000	21.95
70000	22.29
75000	22.42
80000	22.68
85000	22.85
90000	22.93
95000	23.07
100000	23.27
105000	23.31
110000	23.17
115000	23.4
120000	23.39
125000	23.52
130000	23.68
135000	23.72
140000	23.74
145000	23.88
150000	23.85
155000	23.95
160000	23.85
165000	23.94
170000	23.87
175000	23.83
180000	23.92
185000	24.04
190000	24.16
195000	24.17
200000	24.2
205000	24.25
210000	24.24
215000	24.21
220000	24.1
225000	24.2
230000	24.29
235000	24.29
240000	24.34
245000	24.29
250000	24.36
255000	24.48
260000	24.42
265000	24.56
270000	24.6
275000	24.56
280000	24.56
285000	24.57
290000	24.55
295000	24.65
300000	24.75
305000	24.77
310000	24.67
315000	24.78
320000	24.62
325000	24.62
	};
	\addplot[color=blue] table[x=Steps,y=Bleu2] {
Steps	Bleu2
5000	0.09
10000	5.83
15000	14.13
20000	17.2
25000	18.77
30000	19.64
35000	20.05
40000	20.45
45000	20.67
50000	21.19
55000	21.29
60000	21.53
65000	21.63
70000	21.82
75000	22.1
80000	22.33
85000	22.51
90000	22.55
95000	22.66
100000	22.72
105000	22.76
110000	22.89
115000	23.03
120000	23.06
125000	23.26
130000	23.28
135000	23.35
140000	23.33
145000	23.47
150000	23.42
155000	23.39
160000	23.51
165000	23.56
170000	23.59
175000	23.62
180000	23.71
185000	23.65
190000	23.71
195000	23.75
200000	23.81
205000	23.84
210000	23.68
215000	23.61
220000	23.83
225000	23.81
230000	23.94
235000	23.96
240000	23.93
245000	24
250000	23.9
255000	23.84
260000	23.95
265000	24.06
270000	24.19
275000	24.21
280000	24.22
285000	24.13
290000	24.21
295000	24.1
300000	24.09
305000	24.28
310000	24.3
315000	24.29
320000	24.21
325000	24.22
	};
	\legend{BPE, Omorfi+BPE}
	\end{axis}
\end{tikzpicture}
\caption{Evolution of English--Finnish BLEU scores (on $y$-axis) during training.}
\label{fig:en-fi-segm}
\end{figure}

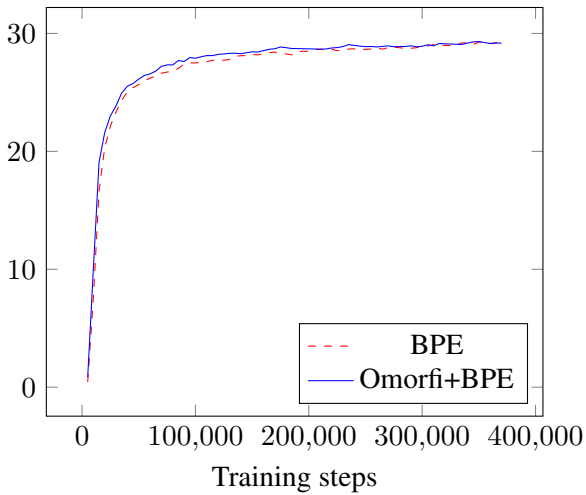
\begin{figure}[h!]
\centering
\begin{tikzpicture}
	\begin{axis}[
	    height=7.0cm,
		xlabel=Training steps,
		x tick label style={/pgf/number format/fixed},
		scaled x ticks=false,
		legend pos=south east
		]
	\addplot[color=red, dashed] table[x=Steps,y=Bleu1] {
Steps	Bleu1
5000	0.43
10000	7.54
15000	16.64
20000	20.31
25000	22.13
30000	23.34
35000	24.33
40000	25.05
45000	25.41
50000	25.66
55000	25.98
60000	26.21
65000	26.37
70000	26.62
75000	26.72
80000	26.8
85000	27.03
90000	27.39
95000	27.49
100000	27.5
105000	27.54
110000	27.59
115000	27.73
120000	27.76
125000	27.72
130000	27.79
135000	28.05
140000	28.11
145000	28.12
150000	28.21
155000	28.18
160000	28.25
165000	28.36
170000	28.42
175000	28.24
180000	28.27
185000	28.18
190000	28.42
195000	28.5
200000	28.46
205000	28.62
210000	28.67
215000	28.61
220000	28.66
225000	28.54
230000	28.58
235000	28.67
240000	28.7
245000	28.63
250000	28.65
255000	28.67
260000	28.76
265000	28.67
270000	28.79
275000	28.83
280000	28.79
285000	28.72
290000	28.74
295000	28.79
300000	28.89
305000	28.94
310000	29.06
315000	29
320000	28.98
325000	28.97
330000	29.06
335000	29.19
340000	29.2
345000	29.05
350000	29.27
355000	29.23
360000	29.13
365000	29.21
370000	29.29
	};
	\addplot[color=blue] table[x=Steps,y=Bleu2] {
Steps	Bleu2
5000	0.86
10000	10.04
15000	19.06
20000	21.59
25000	23
30000	23.86
35000	24.95
40000	25.52
45000	25.75
50000	26.13
55000	26.43
60000	26.57
65000	26.8
70000	27.21
75000	27.33
80000	27.33
85000	27.7
90000	27.62
95000	27.95
100000	27.9
105000	28.02
110000	28.12
115000	28.12
120000	28.22
125000	28.26
130000	28.31
135000	28.32
140000	28.28
145000	28.36
150000	28.44
155000	28.42
160000	28.54
165000	28.65
170000	28.7
175000	28.85
180000	28.79
185000	28.72
190000	28.71
195000	28.7
200000	28.69
205000	28.68
210000	28.66
215000	28.68
220000	28.76
225000	28.8
230000	28.88
235000	29.05
240000	28.98
245000	28.92
250000	28.88
255000	28.89
260000	28.85
265000	28.9
270000	28.95
275000	28.87
280000	28.89
285000	28.89
290000	28.95
295000	28.87
300000	28.91
305000	29.04
310000	28.94
315000	29.14
320000	29.12
325000	29.1
330000	29.08
335000	29.06
340000	29.18
345000	29.28
350000	29.31
355000	29.22
360000	29.15
365000	29.18
370000	29.18
	};
	\legend{BPE, Omorfi+BPE}
	\end{axis}
\end{tikzpicture}
\caption{Evolution of Finnish--English BLEU scores (on $y$-axis) during training.}
\label{fig:fi-en-segm}
\end{figure}

Figures~\ref{fig:en-fi-segm} and~\ref{fig:fi-en-segm} show the evolution of BLEU scores on news2016 during training. For English--Finnish, the Omorfi-segmented system shows slightly higher results during the first 40\,000 training steps, but is then outperformed by the plain BPE-segmented system. For Finnish--English, the Omorfi-segmented system obtains higher BLEU scores much longer, until both systems converge after about 300\,000 training steps.

Table~\ref{tab:segm-results} compares BLEU scores for the 2017 to 2019 test sets. The Omorfi-based system shows consistent improvements when used on the source side, \emph{i.e.} from Finnish to English. However, due to timing constraints, we were not able to integrate the Omorfi-based segmentation into our final submission systems. In any case, the difference observed in the news2019 set after submission deadline is within the bounds of random variation.

\begin{table}[h!]
\centering
\begin{tabular}{lrr}
\toprule
Data set & $\Delta$ BLEU & $\Delta$ BLEU \\
& EN-FI & FI-EN \\
\midrule
news2017 & $-0.47$ & $+0.36$ \\
news2018 & $-0.61$ & $+0.38$ \\
news2019 & $+0.19$ & $+0.04$ \\
\bottomrule
\end{tabular}
\caption{BLEU score differences between Omorfi-segmented and BPE-segmented models. Positive values indicate that the Omorfi+BPE model is better, negative values indicate that the BPE model is better.}
\label{tab:segm-results}
\end{table}

We tested additional transformer models segmented with the SentencePiece toolkit, using a shared vocabulary of 40k tokens trained only on the parallel corpora. We do this with the purpose of comparing the use of a software tailored specifically for Finnish language (Omorfi) with a more general segmentation one. These models were trained with the same specifications as the previous ones, including the transformer hyperparameters, the train and development data and the domain-labeling. Since  we used OpenNMT-py to train these models, it is difficult to know whether the differences come from the segmentation or the toolkit. We, however, find it informative to present these results. Table~\ref{tab:spm} presents the obtained BLEU scores with both systems.

We notice that both systems yield similar scores for both translation directions. SentencePiece models are consistently ahead of Omorfi+BPE, but this difference is so small that it cannot be considered convincing nor significant.
\begin{table}[h!]
\centering
\begin{tabular}{lccc}
\toprule
Model & &news& news \\
 & &2017& 2019 \\
\midrule
SentencePiece& EN-FI   & $25.60$ &	$20.60$ \\
Omorfi+BPE   &EN-FI    & $25.50$ & $20.13$ \\
SentencePiece& EN-FI   & $31.50$ & 	$25.00$ \\
Omorfi+BPE   &FI-EN    & $31.21 $ &	$24.06$ \\
\bottomrule
\end{tabular}
\caption{BLEU scores comparison between SentencePiece and  Omorfi+BPE-segmented models.}
\label{tab:spm}
\end{table}

Our final models for English-to-Finnish are standard transformer models with BPE-based segmentation, trained using MarianNMT with the same settings and hyper-parameters as the other experiments. We used the filtered training data using the relaxed settings of the language model filter to obtain better coverage for this language pair. The provided training data is much smaller and we also have less back-translated data at our disposal, which motivated us to lower the threshold of taking examples from web-crawled data. Domain fine-tuning is done as well using news test sets from 2015, 2016 and 2018. The results on development test data from 2017 are listed in Table~\ref{tab:marian_enfi}.

\begin{table}[h!]
    \centering
    \begin{tabular}{lcc}
    \toprule
            & \multicolumn{2}{c}{BLEU news2017} \\
    Model   & L2R & R2L \\
    \midrule
    Run 1    & 27.68 & 28.01 \\
    Run 2    & 28.64 & 28.77 \\
    Run 3    & 28.64 & 28.41 \\
    \midrule
    Ensemble  & 29.54 & 29.76\\
    Rescored  & 29.60 & 29.72\\
    -- L2R+R2L & \multicolumn{2}{c}{\bf 30.66} \\
    \midrule
    Top matrix & \multicolumn{2}{c}{21.7} \\
    \bottomrule
    \end{tabular}
    \caption{Results from individual MarianNMT transformer models and their combinations for English to Finnish (cased BLEU). The {\em top matrix} result refers to the best system reported in the on-line evaluation matrix (accessed on May 16, 2019).}
    \label{tab:marian_enfi}
\end{table}

A striking difference to English--German is that right-to-left decoding models are on par with the other direction. The scores are substantially higher than the currently best (post-WMT 2017) system reported in the on-line evaluation matrix for this test set, even though this also refers to a transformer with a similar architecture and back-translated monolingual data. This system does not contain data derived from ParaCrawl, which was not available at the time, and the improvements we achieve demonstrate the effectiveness of our data filtering techniques from the noisy on-line data.

For Finnish-to-English, we trained MarianNMT models using the same transformer architecture as for the other language pairs. Table~\ref{tab:marian_fien} shows the scores of individual models and their combinations on the development test set of news from WMT 2017. All models are trained on the same filtered training data using the strict settings of the language model filter including the back-translations produced for English monolingual news.

\begin{table}[h!]
    \centering
    \begin{tabular}{lcc}
    \toprule
            & \multicolumn{2}{c}{BLEU news2017} \\
    Model   & L2R & R2L \\
    \midrule
    Run 1    & 32.26 & 31.70 \\
    Run 2    & 31.91 & 31.83 \\
    Run 3    & 32.68 & 31.81 \\
    \midrule
    Ensemble  & 33.23 & 33.03\\
    Rescored  & 33.34 & 32.98\\
    -- L2R+R2L & \multicolumn{2}{c}{\bf 33.95} \\
    \midrule
    Top (with ParaCrawl) & \multicolumn{2}{c}{34.6} \\
    Top (without ParaCrawl) & \multicolumn{2}{c}{25.9} \\
    \bottomrule

    \end{tabular}
    \caption{Results from individual MarianNMT transformer models and their combinations for Finnish to English (cased BLEU). Results denoted as top refer to the top systems reported at the on-line evaluation matrix (accessed on May 16, 2019), one trained with the 2019 data sets and one with 2017 data.}
    \label{tab:marian_fien}
\end{table}

In contrast to English-to-German, models in the two decoding directions are quite similar again and the difference between left-to-right and right-to-left models is rather small. 
The importance of the new data sets from 2019 are visible again and our system performs similarly, but still slightly below the best system that has been submitted this year to the on-line evaluation matrix on the 2017 test set.

\subsection{The English--Finnish rule-based system} \label{sec:rbmt}

Since the WMT 2018 challenge, there has been development in four areas of translation process in the rule-based system for English--Finnish:

\begin{enumerate}
\item The standard method in handling English noun compounds was to treat them as multiword expressions (MWE). This method allows many kinds of translations, even multiple translation, which can be handled in semantic disambiguation. However, because noun compounding is a common phenomenon, also a default handling method was developed for such cases, where two or more consecutive nouns are individually translated and glued together as a single word. The system works so that if the noun combination is not handled as MWE, the second strategy is applied~\cite{Hurskainen2018a}.
\item The translation of various types of questions has been improved. Especially the translation of indirect questions was defective, because the use of\emph{if} in the role of initiating the indirect question was not implemented. The conjunction\emph{if} is ambiguous, because it is used also for initiating the conditional clause~\cite{Hurskainen2018b}.
\item Substantial rule optimizing was carried out. When rules are added in development process, the result is often not optimal. There are obsolete rules and the rules may need new ordering. As a result, a substantial number of rules (30\%) were removed and others were reordered. This has effect on translation speed but not on translation result~\cite{Hurskainen2018c}.
\item Temporal subordinate clauses, which start with the conjunction \emph{when} or \emph{while}, can be translated with corresponding subordinate clauses in Finnish. However, such clauses are often translated with participial phrase constructions. Translation with such constructions was tested. The results show that although they can be implemented, they are prone to mistakes~\cite{Hurskainen2018d}.
\end{enumerate}

These improvements to the translation system contribute to fluency and accuracy of translations.

\section{Document-level approaches} \label{sec:doclevel}

To evaluate the effectiveness of various document-level translation approaches for the English--German language pair, we experimented with a number of different approaches which are described below. In order to test the ability of the system to pick up document-level information, we also created a shuffled version of the news data from 2018.
We then test our systems on both the original test set with coherent test data divided into short news documents and the shuffled test set with broken coherence.

\subsection{Concatenation models}

Some of the previously published approaches use concatenation of multiple source-side sentences in order to extend the context of the currently translated sentence~\citep{Tiedemann2017}. In addition to the source-side concatenation model, we also tested an approach where we concatenate the previously translated sentence with the current source sentence. The concatenation approaches we tested are listed below.
\begin{itemize}
    \item MT-concat-source: (2+1) Concatenating previous source sentence with the current source sentence~\citep{Tiedemann2017}. (3+1a) Concatenating the previous two sentences with the current source sentence. (3+1b) Concatenating the previous, the current and the next sentence in the source languages.
    \item MT-concat-target: (1t+1s+1) Concatenating the previously translated (target) sentence with the current source sentence.
    \item MT-concat-source-target: (2+2) Concatenating the previous with the current source sentence and translate into the previous and the current target sentence~\citep{Tiedemann2017}. Only the second sentence in the translation will be kept for evaluation of the translation quality.
\end{itemize}


Extended context models only make sense with coherent training data. Therefore, we ran experiments only with the training data that contain translated documents, \emph{i.e.} Europarl, NewsCommentary, Rapid and the back-translations of the German news from 2018. Hence, the baseline is lower than a sentence-level model on the complete data sets provided by WMT. Table~\ref{tab:concat-models} summarizes the results on the development test data (news 2018).

\begin{table}[h!]
    \centering
    \begin{tabular}{lcc}
    \toprule
                 & \multicolumn{2}{c}{BLEU news2018} \\
System           & Shuffled & Coherent \\
\midrule
Baseline         & 38.96    & 38.96 \\
2+1              & 36.62    & 37.17 \\
3+1a             & 33.90    & 34.30 \\
3+1b             & 34.14    & 34.39 \\
1t+1s+1          & 36.82    & 37.24 \\
2+2              & 38.53    & {\bf 39.08} \\
\bottomrule
    \end{tabular}
    \caption{Comparison of concatenation approaches for English--German document-level translation.}
    \label{tab:concat-models}
\end{table}

The results overall are rather disappointing. All but one of the concatenation models underperform and cannot beat the sentence-level baseline. Note that the concat-target model (1t+1s+1) even refers to an oracle experiment in which the reference translation of the previous sentence is fed into the translation model for translating the current source sentence. As this is not very successful, we did not even try to run a proper evaluation with system output provided as target context during testing. Besides the shortcomings, we can nevertheless see a consistent pattern that the extended context models indeed pick up information from discourse. For all models we observe a gain of about half a BLEU point when comparing the shuffled to the non-shuffled versions of the test set. This is interesting and encourages us to study these models further in future work, possibly with different data sets, training procedures and slightly different architectures.


\subsection{Hierarchical attention models}

A number of approaches have been developed to utilize the attention mechanism to capture extended context for document-level translation. We experimented with the two following models:

\begin{itemize}
    \item NMT-HAN: Sentence-level transformer model with a hierarchical attention network to capture the document-level context~\citep{Miculicich_EMNLP_2018}.
    \item selectAttn: Selective attention model for context-aware neural machine translation~\citep{maruf-etal-2019-selective}.
\end{itemize}


For testing the selectAttn model, we used the same data with document-level information as we applied in the concatenation models. For NMT-HAN we had to use a smaller training set due to lack of resources 
and due to the implementation not supporting data shards.
For NMT-HAN we used only Europarl, NewsCommentary and Rapid for training. Table~\ref{tab:HAN-models} summarizes the results on the development test data. Both of the tested models need to be trained on sentence-level first, before tuning the document-level components. 

\begin{table}[h!]
    \centering
    \begin{tabular}{lcc}
\toprule
Model          &  Sentence-level & Document-level\\
\midrule
NMT-HAN            &  35.03          & 31.73 \\
selectAttn     &  35.26          & 34.75 \\ 
\bottomrule
    \end{tabular}
    \caption{Results (case-sensitive BLEU) of the hierarchical attention models on the coherent newstest 2018 dataset.}
    \label{tab:HAN-models}
\end{table}


The architecture of the selective attention model is based on the general transformer model but with quite a different setup in terms of hyperparameters and dimensions of layer components etc. We applied the basic settings following the documentation of the software. In particular, the model includes 4 layers and 8 attention heads, and the dimensionality of the hidden layers is 512. We applied a sublayer and attention dropout of 0.1 and trained the sentence-level model for about 3.5 epochs. We selected monolingual source-side context for our experiments and hierarchical document attention with sparse softmax. Otherwise, we also apply the default parameters suggested in the documentation with respect to optimizers, learning rates and dropout. Unfortunately, the results do not look very promising as we can see in Table~\ref{tab:HAN-models}. The document-level model does not even reach the performance of the sentence-level model even though we trained until convergence on development data with patience of 10 reporting steps, which is quite disappointing. Overall, the scores are below the standard transformer models of the other experiments, and hence, we did not try to further optimize the results using that model. 

For the NMT-HAN model we used the implementation of~\citet{Miculicich_EMNLP_2018} with the recommended hyperparameter values and settings. The system is based on the OpenNMT-py implementation of the transformer. The model includes 6 hidden layers on both the encoder and decoder side with a dimensionality of 512 and the multi-head attention has 8 attention heads. We applied a sublayer and attention dropout of 0.1. The target and source vocabulary size is 30K. We trained the sentence-level model for 20 epochs after which we further fine-tuned the encoder side hierarchical attention for 1 epoch and the joint encoder-decoder hierarchical attention for 1 epoch. The results for the NMT-HAN model are disappointing. The document-level model performs significantly worse than the sentence-level model.

\section{Results from WMT 2019} \label{sec:finalsystems}

Table~\ref{tab:results2019} summarizes our results from the WMT 2019 news task. We list the official score from the submitted systems and post-WMT scores that come from models described above. For Finnish--English and English--Finnish, the submitted systems correspond to premature single models that did not converge yet. Our submitted English--German model is the ensemble of 9 models described in Section~\ref{sec:en_de}. 

\begin{table}[h!]
    \centering
    \begin{tabular}{llr}
    \toprule
    Language pair         & Model     & BLEU \\
    \midrule
    English--German & submitted & 41.4 \\
                      & L2R+R2L   & 42.95 \\
    Finnish--English & submitted & 26.7\\
                       & L2R+R2L & 27.80\\
    English--Finnish & submitted & 20.8\\
                       & rule-based & 8.9\\
                       & L2R+R2L   & 23.4 \\
    \bottomrule
    \end{tabular}
    \caption{Final results (case-sensitive BLEU scores) on the 2019 news test set; partially obtained after the deadline.}
    \label{tab:results2019}
\end{table}

The ensemble results clearly outperform those results but were not ready in time. We are still below the best performing system from the official participants of this year's campaign but the final models perform in the top-range of all the three tasks. For English--Finnish, our final score would end up on a third place (12 submissions from 8 participants), for Finnish--English it would be the fourth-best participant (out of 9), and English--German fifth-best participant (out of 19 with 28 submissions).

\section{Conclusions}
In this paper, we presented our submission for the WMT 2019 news translation task in three language pairs: English--German, English--Finnish and Finnish--English.

For all the language pairs we spent considerable time on cleaning and filtering the training data, which resulted in a significant reduction of training examples without a negative impact on translation quality.

For English--German we focused both on sentence-level neural machine translation models as well as document-level models. For English--Finnish, our submissions consists of an NMT system as well as a rule-based system whereas the Finnish--English system is an NMT system. For the English--Finnish and Finnish--English language pairs, we compared the impact of different segmentation approaches. Our results show that the different segmentation approaches do not significantly impact BLEU scores. However, our experiments highlight the well-known fact that ensembling and domain adaptation have a significant positive impact on translation quality.

One surprising finding was that none of the document-level approaches really worked, with some even having a negative effect on translation quality.

\section*{Acknowledgments}
\vspace{1ex}
\noindent
\begin{minipage}{0.1\linewidth}
   \raisebox{-0.2\height}{\includegraphics[trim =32mm 55mm 30mm 5mm, clip, scale=0.2]{erc.ai}}
\end{minipage}
\hspace{0.01\linewidth}
\begin{minipage}{0.70\linewidth}The work in this paper was supported by the FoTran project, funded by the European Research Council (ERC) under the European Union’s Horizon 2020 research and innovation programme (grant agreement No 771113), and the MeMAD project funded by the European Union’s Horizon 2020 Research and Innovation Programme under grant agreement No 780069.

 \vspace{1ex}
\end{minipage}
\hspace{0.01\linewidth}
\begin{minipage}{0.05\linewidth}
 \vspace{0.05cm}
\raisebox{-0.25\height}{\includegraphics[trim =0mm 5mm 5mm 2mm,clip,scale=0.078]{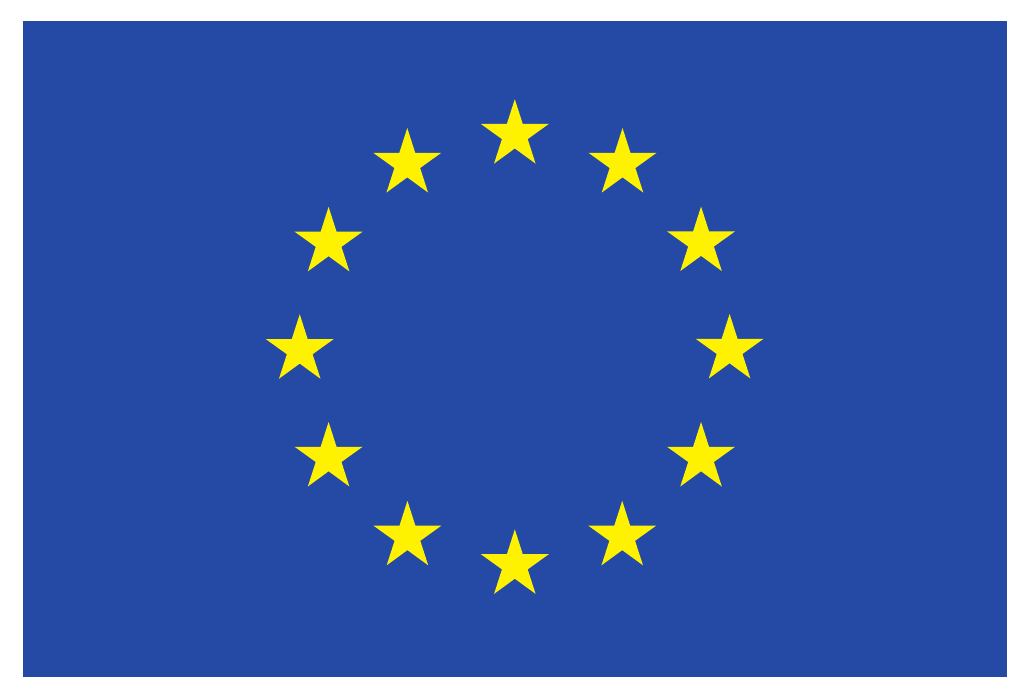}}
\vspace{0.05cm}

\end{minipage}



The authors gratefully acknowledge the support of the Academy of Finland through project 314062 from the ICT 2023 call on Computation, Machine Learning and Artificial Intelligence and project 270354/273457.
\appendix

\bibliography{references}
\bibliographystyle{acl_natbib}

\end{document}